\documentclass[letterpaper, 10 pt, conference]{ieeeconf}
\IEEEoverridecommandlockouts

\usepackage[colorlinks]{hyperref}
\hypersetup{
    colorlinks,
    linkcolor=black,
    citecolor=black,
    filecolor=magenta,
    urlcolor=cyan,
}

\usepackage{resizegather}

\usepackage{cite}
\usepackage{amsmath,amssymb,amsfonts,mathrsfs}
\usepackage{graphicx}
\usepackage{textcomp}
\usepackage{xcolor}
\usepackage{tcolorbox}
\usepackage{textcomp}
\usepackage{tablefootnote}
\usepackage{threeparttable}
\usepackage{caption, subcaption}
\usepackage{bbm}
\usepackage{dsfont}
\usepackage{tikz}
\usepackage{graphicx}
\usepackage{tkz-euclide}
\usepackage{algorithm}
\usepackage{algpseudocode}
\usetikzlibrary{positioning}
\usetikzlibrary{shapes}
\usetikzlibrary{shapes.misc}
\usetikzlibrary{shapes.geometric}
\usetikzlibrary{plotmarks}
\usetikzlibrary{intersections}
\usetikzlibrary{calc}
\usetikzlibrary{fit}
\usetikzlibrary{patterns,tikzmark}
\usetikzlibrary{matrix,decorations.pathreplacing,calc}

\tikzset{cross/.style={cross out, draw, 
         minimum size=2*(#1-\pgflinewidth), 
         inner sep=0pt, outer sep=0pt}}
\usepackage{amsmath,amssymb,amsfonts,mathrsfs}
\usepackage{cleveref}
\newcommand{\vertiii}[1]{{\left\vert\kern-0.25ex\left\vert\kern-0.25ex\left\vert #1 
    \right\vert\kern-0.25ex\right\vert\kern-0.25ex\right\vert}}

\makeatletter
\renewcommand{\fps@figure}{htp}
\renewcommand{\fps@table}{htp}
\makeatother
\usepackage{booktabs}

\def\BibTeX{{\rm B\kern-.05em{\sc i\kern-.025em b}\kern-.08em
    T\kern-.1667em\lower.7ex\hbox{E}\kern-.125emX}}

\begin{document}

\title{Value function based Reward shaping for safety in quadrupeds}
% Some Random Title can be changed later

\author{Janani S K$^{1}$, Manan Tayal$^{2}$, and  Shishir Kolathaya$^{2}$% % <-this % stops a space
\thanks{
}
\thanks{$^{1}$Department of Engineering Design, Indian Institute of Technology, Madras.
{\tt\scriptsize ed22b017@smail.iitm.ac.in}
.
}% <-this % stops a space
\thanks{$^{2}$ Center for Cyber-Physical Systems, Indian Institute of Science (IISc), Bengaluru.
{\tt\scriptsize \{manantayal, shishirk\}@iisc.ac.in}
.
}%
}

\newcommand{\mcom}[1]{{\color{red}{{[MT: #1]}}}}
\newcommand{\jcom}[1]{{\color{blue}{{[J: #1]}}}}
\maketitle
\begin{abstract}
Reinforcement Learning (RL) has progressed
from simple control tasks to complex real-world challenges with
large state spaces. While RL excels in these tasks, training time
remains a limitation. Reward shaping is a popular solution,
but existing methods often rely on value functions, which face
scalability issues. This paper presents a novel safety-oriented
reward-shaping framework inspired by barrier functions, offering simplicity and ease of implementation across various
environments and tasks. To evaluate the effectiveness of the proposed reward formulations, we conduct simulation experiments on the Unitree Go1 quadruped robot in Isaacgym. The policy was better than the existing baselines and reduced safety violations from $27\%$ to $0.1\%$ without almost any compromise on the performance. 

\end{abstract}

\section{Introduction}
\label{section: Introduction}

Reinforcement learning (RL) has moved from simple control benchmarks to high-dimensional, real-world problems, and legged locomotion is one of its clearest successes. Policies trained in simulation now let quadrupeds walk, run and climb across varied terrain, including stairs and uneven ground~\cite{long2024hybridinternalmodellearning}. In these settings the reward function does most of the work. It specifies task objectives such as velocity tracking, and also regularisers such as joint power, action rate and smoothness. The resulting policies are optimised for expected return, but nothing in the reward guarantees that the robot stays within its safe operating limits. On hardware, joint angles that exceed their limits can damage the actuators and the robot.

A common way to enforce such limits is to clip the actions or joint targets. Clipping is simple, but it acts only at the output of the policy and gives the learner no signal about \emph{how close} it is to the boundary and also it doesn't guarantee that performance will be maintained. The policy therefore keeps operating at the edge of the safe set, and clipping alone produced safety violations in $27\%$ of our test steps. Safety can instead be built into the reward through reward shaping. Many shaping schemes depend on learned value functions, which are hard to scale to high-dimensional state spaces and make the shaping term difficult to implement and tune.

Control barrier functions (CBFs) offer a principled alternative. A CBF defines a safe set as the superlevel set of a function $l(x)$ and requires $\dot{l}(x) + \kappa(l(x)) \ge 0$, which makes the set forward invariant. This condition can be turned directly into a reward term. Hand-designed CBFs have two weaknesses. They are \emph{reactive}, so they only guarantee safety one step ahead and do not account for control-input constraints.Using CBFs in the QP formulation completely overwrites the performance controller and thus there is no guarantee on the performance of the system. They also need the system dynamics, which are rarely available in closed form for legged robots. Hamilton--Jacobi (HJ) reachability addresses the first weakness: the safe value function it computes is a valid barrier function that accounts for future behavior under input constraints~\cite{9561949}. However, for solving the HJ equation still system dynamics is needed. Recently, value-guided offline control barrier functions (V-OCBF)~\cite{tayal2026vocbflearningsafetyfilters} learn this kind of safe value function directly from offline transition data $\{x_t, u_t, x_{t+1}\}$ and the safe set function $l(x)$ with no knowledge of the dynamics.

In this paper we bring these ideas together into a safety-oriented reward-shaping framework for quadruped locomotion. We consider box constraints of the form $x_{\min} \le x \le x_{\max}$ on the joint angles. We first construct a smooth barrier function for each joint and then learn a V-OCBF safe value function $B_{\psi}^{j}$ for each joint from data collected with random actions. The per-joint values are combined with a softmin, which keeps the result differentiable, and the final barrier $B_\psi(s)$ is used to shape the reward through
\begin{equation}
    r^{\mathrm{VOCBF}}(s,\dot{s}) = \dot{B}_\psi(s) + \gamma B_\psi(s).
\end{equation}
The state derivative is estimated by finite differences from the simulator, so the dynamics are never required.

The rest of the paper is organised as follows. Section~\ref{section: Background} reviews barrier functions, HJ reachability and V-OCBF. Section~\ref{section: Method} presents the barrier construction and the VOCBF reward shaping. Section~\ref{section: Simulation} describes the experiments and results, and Section~\ref{section: Conclusions} concludes.

% ---------------------------------------------------------------
% Suggested labels for the other sections:
%   \section{Preliminaries}\label{sec:prelim}
%   \section{Methodology}\label{sec:method}
%   \section{Simulations}\label{sec:sim}
%   \section{Conclusions}\label{sec:concl}
%
% Suggested BibTeX keys (match to your .bib file):
%   bansal2021deepreach  -> reference [1]
%   tayal2026vocbf       -> reference [2]
%   long2024hybrid       -> reference [3]
%   TODO_reward_shaping  -> add a reward-shaping reference
% ---------------------------------------------------------------
\section{Preliminaries}
\label{section: Background}
In this section we formally introduce Barrier functions, HJ Reachability based CBF, Value guided offline Control barrier functions, 

\subsection{Barrier functions}
Consider the safe set $\mathcal{C}$ which is a \textit{Super level set} of a function $l(x)$.
\begin{align*}
    \mathcal{C}=& \{x \in \mathcal{S} : l(x) \geq 0\}\\
    \text{Int}(\mathcal{C})=& \{ x \in \mathcal{S} : l(x) >0\}\\
    \partial\mathcal{C}=&\{x \in \mathcal{S} : l(x)=0\}
\end{align*}
We want to render the set $\mathcal{C}$ forward invariant, that is , $x(0) \in \mathcal{C} \implies x(t) \in \mathcal{C} \quad \forall t \geq 0 $. 
\textit{Given set $\mathcal{C}$, the function $l$ is called Barrier function if there exists an extended class $\mathcal{K}$ function $\kappa$ such that $\forall s \in \mathcal{S}$:}
\begin{align*}
    \dot{l}(s,\dot{s}) +\kappa(l(s)) \geq 0 
\end{align*}
For a barrier function $l$, $\dot{l}(s,\dot{s})=[\frac{dl}{ds}]^T\dot{s}$. Even if the dynamics of the system is not known explicitly, $\dot{s}$ is calculated using $\dot{s}=\frac{s_t-s_{t-1}}{\Delta t}$.
\subsection{HJ Reachability based CBF}
The CBFs mentioned above are reactive- only guarantee one step safety and they don't account for control input constraints. $\zeta^{u,d}_{x,t}$ is the rollout of the state with inputs given by $u$ and disturbance $d$ from $t$ to $T$ stating at $x$ with the dynamics given by $\dot{x}=f(x,u,d)$, $\zeta^{u,d}_{x,t}(\tau)$ is the state $x_\tau$ at time $\tau$ in the rollout $\zeta^{u,d}_{x,t}$.[Refer \cite{9561949}]
\begin{align*}
    J(x,t,u(.),d(.))=\underset{\tau \in [t,T]}{min}\{l(\zeta^{u,d}_{x,t}(\tau))\}\\
    V_t(x)=\underset{u_{t:T}\in U_{t:T}}{max}\{ \underset{d_{t:T}\in D_{t:T}}{min}\{J(x,t,u,d)\}\}
\end{align*}
Using this, the following statements can be made:
\begin{align*}
V_t(x) \geq 0 \iff{}& \exists\, u_{t:T} \in U_{t:T} \text{ s.t. } l(x_\tau) \geq 0 \\
& \forall\, d_{t:T} \in D_{t:T},\ \tau \in [t, T], \\[4pt]
V_t(x) < 0 \iff{}& \forall\, u_{t:T} \in U_{t:T},\ \exists\, \tau \in [t, T],\ d_{t:T} \in D_{t:T} \\
& \text{s.t. } l(x_\tau) < 0.
\end{align*}
$V_t(x)$ is safe value function. Finding such $V_t(x)$ and maximizer $u_{t:T}$.
$V_t(x)$ is found by solving the Hamilton-Jacobi-Isaacs Variational Inequality (HJI VI):
\begin{align*}
    min \{D_tV_t(x)+ H_t(x),l(x)-V_t(x)\}=0\\
    H_t(x)=\underset{u}{max}\quad\underset{d}{min} \langle\nabla V_t(x), f(x,u,d)\rangle. 
\end{align*}
$V_t(x)$ can be used as a barrier function for the same set $\mathcal{C}$.
\subsection{Value guided offline control barrier functions}
To get the safe value from demonstrations, \cite{tayal2026vocbflearningsafetyfilters} provides a method to get the value function based barrier function based on demonstrations even for systems where the dynamics is not known. From the demonstration dataset $\mathcal{D}$ with trajectories in the form $\{x_t,u_t,x_{t+1}\}^N_{t=0}$, barrier update is given by:
\begin{align*}
    B(x_t)=\min\Bigl\{l(x_t),\, \max_{u_t}\,B(x_{t+1})\Bigr\}
    \quad \forall t \in \{0,1,2, \cdots , N\}
\end{align*}
The barrier function is approximated using 2 neural networks with the following loss functions:
\begin{align*}
L(\theta) ={}& \mathbb{E}_{(x_t,u_t,x_{t+1})\sim\mathcal{D}}\Big[\big((1-\gamma)\,l(x_t) \\
&\qquad +\gamma \min\{l(x_t),B_\theta(x_{t+1})\}-B_\theta(x_t)\big)^2\Big], \\[4pt]
L(\psi) ={}& \mathbb{E}_{(x_t,u_t,x_{t+1})\sim\mathcal{D}}\Big[\big|\tau-\mathbf{1}_{B_\theta(x_t)-B_\psi(x_t)\le 0}\big| \\
&\qquad \cdot \big(B_\theta(x_t)-B_\psi(x_t)\big)^2\Big].
\end{align*}
where $\gamma, \tau$ are hyper parameters. $\gamma$ is used to avoid getting trivial solutions and the loss function corresponding to $\psi$ is expectile regression loss function, keeping $\tau>0.5$ penalizes the cases where $B_\psi(x_t)<B_\theta(x_t)$ than the other case and thus approximating $max_{u_t} B_\theta(x_{t+1})$ while keeping the control $u_t$ within the limits. The safe barrier function used is the learned $B_\psi(x_t)$. Note that there is no need for dynamics to learn this safe value function.

\label{section: Method}
\section{METHODOLOGY}
The safe set of interest is $s^l_{min} \leq s^l \leq s^l_{max}$ where $s^l$ is $l^{th}$ element in the state vector. We first investigate the barrier functions that can be used for these constraints and then how reward shaping is done using VOCBF.
\subsection{Barrier function}
\begin{figure}
    \centering
    \includegraphics[width=0.8\linewidth]{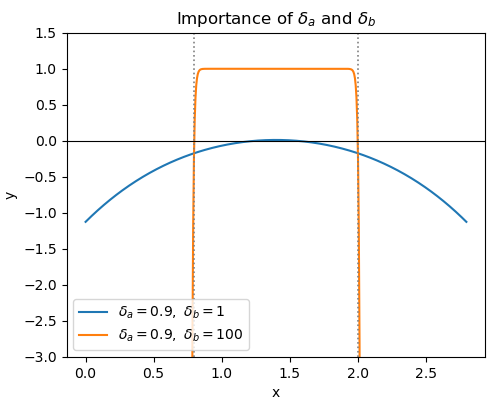}
    \caption{Importance of $\delta_a, \delta_b$}
    \label{fig:importance}
\end{figure}

The set $\mathcal{C}=\{s \in \mathcal{S} \subseteq \mathbb{R}^n :s_{min} \leq s \leq s_{max}\}$ can be written as a \textit{super level set} of function $l$ with appropriate hyperparameters $\delta_a, \delta_b$ such that:
\begin{align*}
    l(s,\delta)=\underset{i}{min}\{ 1-\delta^i_a(e^{\delta^i_b(s^i-s^i_{max})}+e^{\delta^i_b(s^i_{min}-s^i)})\} 
\end{align*}
\textit{Choosing appropriate $\delta_a^i,\delta_b^i$}:
\newline
For each $l^i(s^i,\delta^i)$, the following conditions should be satisfied:
\begin{align*}
    l^i(s^i,\delta^i)& >0 \quad \forall \quad s^i_{min} <s^i< s^i_{max}\\
    l^i(s^i,\delta^i)& =0  \text{  when  } s^i=s^i_{min} \text{  and  } s ^i=s^i_{max} \\
    l^i(s^i,\delta^i)& < 0  \quad \forall  s^i: s^i> s^i_{max} \text{  or  } s^i < s^i_{min}
 \end{align*} \label{eq:barrier}
Using these conditions, we find the conditions that $\delta^i$ must satisfy for $\mathcal{C}^i$ to be represented as \textit{super level set }of $l^i(s^i,\delta^i)$. 
\begin{align*}
    l^i(s^i,\delta^i)&=1-\delta^i_a(e^{\delta^i_b(s^i-s^i_{max})}+e^{\delta^i_b(s^i_{min}-s^i)})\\
    & =1-pe^{\delta^i_bs^i}-qe^{-\delta^i_bs^i} , p=\delta^i_ae^{-\delta^i_bs^i_{max}} , q =\delta^i_ae^{\delta^i_bs^i_{min}}\\
\end{align*}
Just from the definition of $l^i(s^i,\delta^i)$, $\delta^i_a>0$ else $l^i(s^i,\delta^i)>0 \forall s^i$, this is the first condition.
\newline
Taking $u=e^{\delta^i_bs^i}$ and solving for the roots of the equation $l^i(u,\delta^i)=0$, 
\begin{align*}
    l^i(u,\delta^i)&=1-pu-\frac{q}{u}=0\\
    u-&pu^2-q=0 \\
    r_1=&\frac{1 - \sqrt{1-4pq}}{2p}\\
    r_2=&\frac{1 + \sqrt{1-4pq}}{2p}\\
\end{align*}
From this clearly, $u-pu^2-q \geq0 $ when $r_1\leq u \leq r_2$ and otherwise when $u <r_1, u> r_2$. Since $u=e^{\delta^i_bs^i},u >0 \quad \forall s^i \in \mathbb{R}$, so sign of $l^i(u,\delta^i)$ is same as that of $u-pu^2-q$. For $l^i(u,\delta^i)=0 $ at $s^i=s^i_{min} $and $ s^i_{max}$, $\delta^i$ should satisfy the condition, $e^{\delta^i_bs^i_{min}}=r_1$ and $e^{\delta^i_bs^i_{max}=}r_2$ as $\delta^i_a>0, $ so $ r_1\leq r_2$. On substituting for $p,q$, the final conditions are $e^{2\delta^i_b(s^i_{min}-s^i_{max})}=\frac{1-\delta^i_a}{\delta^i_a}, \delta^i_a>0$ Only these $\delta^i$, satisfy \cref{eq:barrier} and vice-versa. From this $\delta_a \in (0,1]$. In Fig.\ref{fig:importance} , set $0.8\leq x \leq 2$ is safe set, it can be seen that when $\delta_a=0.9, \delta_b=1$, the function $l(x)<0 \quad \forall x$, but on using $\delta_a=0.9, \delta_b=100$, which satisfies the condition provides a valid barrier function.

From here on, $\delta^i$ is dropped in $l^i(s^i,\delta^i)$ and all the $l^i(s^i)$ satisfy the required conditions.
Now, $l(s)=\underset{i}{min}\{l^i(s^i)\}$, satisfies the following conditions:
\begin{align*}
    l(s)&\geq0 \iff l^i(s^i) \geq 0 \quad \forall i \iff s \in \mathcal{C}\\
    l(s) &<0 \iff \exists i : l^i(s^i)<0 \iff s \notin \mathcal{C}
\end{align*}
Since $l(s)$ is not differentiable, we use softmin instead of min which is denoted by $l_{sm}(s,\beta)={softmin_\beta}(\{l^i(s^i)\})$.
Softmin of vector $x$ with hyperparameter $\beta>0$ is defined by:
\begin{align*}
    softmin_\beta(x)=-\frac{1}{\beta}ln(\sum_i e^{-\beta x_i})
\end{align*}
Accuracy of softmin approximating the min function depends on $\beta$, large $\beta$ is more accurate where as smaller ones are more smooth. Softmin function is a differentiable function, which solves the problem with using min.
\subsection{Reward shaping with VOCBF}
Using the above mentioned $l^j(s)$, safe value function $B^j_{\psi^j}(s^j)$ is  using VOCBF method is learnt for each joint and then softmin of $B^j_{\psi^j}(s^j)$ is taken, and then using that:
\begin{align*}
    B_\psi(s)=&softmin_\beta([B^j_{\psi^j}(s^j)]_j)\\
    r^{VOCBF}(s,\dot{s})=&\dot{B_\psi}(s)+\gamma B_\psi(s)=[\frac{\partial B_\psi(s)}{\partial s}]^T\dot{s}+ \gamma B_{\psi}(s)
\end{align*}
For this, the activation function that is used in the neural network to approximate $B_\psi$ should be differentiable, i.e activation functions such as tanh, linear etc can be used, but others such as ReLU cannot be used.

Even if the dynamics is not known i.e $\dot{s}$, it can be found using $\dot{s}=\frac{s_t-s_{t-1}}{\Delta t}$ from the simulation.

\section{Simulations}
\label{section: Simulation}
The experiments were done for the Unitree Go-1 robot in the Isaacgym simulator. The rewards same that of HIMLOCO \cite{long2024hybridinternalmodellearning} and on top of that the safety rewards were added. The method used to train the RL policy is also same as that of HIMLOCO. 

Training the VOCBF: The dataset for training VOCBF $\mathcal{D}=\{x_t,u_t,x_{t+1}\}_{t=0}^N$ is generated by using random $u_t$ at $x_t$. For each $B^j_{\psi^j}$, the parameters $\gamma=0.99, \tau =0.8$ are used to train. The best $\beta$ for the softmin function was 40. All the policies were trained for $1680$ episodes.
\begin{figure}
    \centering
    \includegraphics[width=0.5\linewidth]{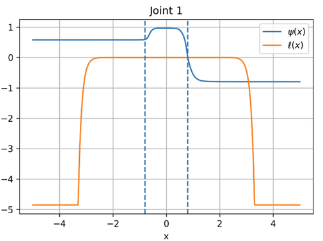}
    \caption{Learnt barrier function and barrier function for joint-1 of unitree go-1}
    \label{fig:barrier plot}
\end{figure}
\begin{figure*}[t]
    \centering
    \fbox{%
    \begin{minipage}{\dimexpr\textwidth-2\fboxsep-2\fboxrule\relax}
        \centering
        \begin{subfigure}[t]{0.49\linewidth}
            \centering
            \includegraphics[width=\linewidth, height=6cm, keepaspectratio=false]{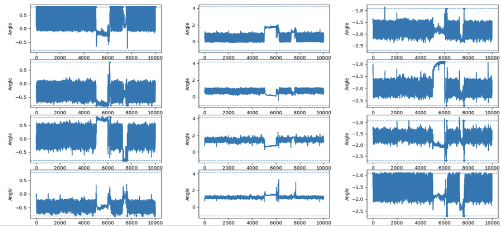}
            \caption{Only clipping}
            \label{fig:only_clipping}
        \end{subfigure}\hfill
        \begin{subfigure}[t]{0.49\linewidth}
            \centering
            \includegraphics[width=\linewidth, height=6cm, keepaspectratio=false]{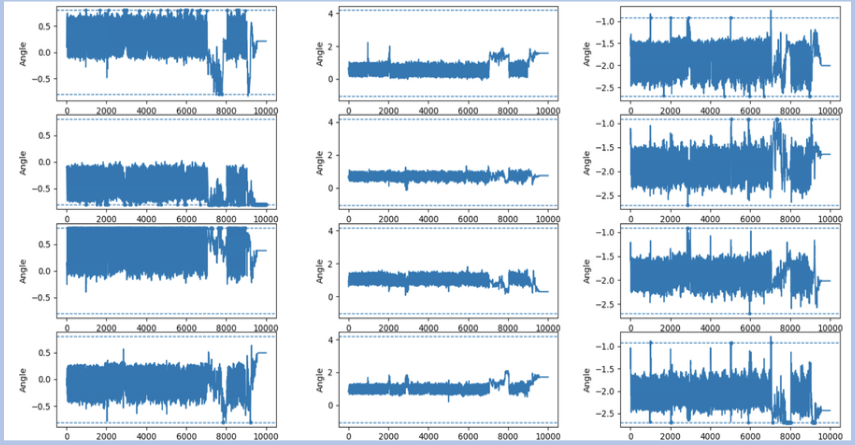}
            \caption{Sum CBF}
            \label{fig:sum_cbf}
        \end{subfigure}

        \vspace{3mm}

        \begin{subfigure}[t]{0.49\linewidth}
            \centering
            \includegraphics[width=\linewidth, height=6cm, keepaspectratio=false]{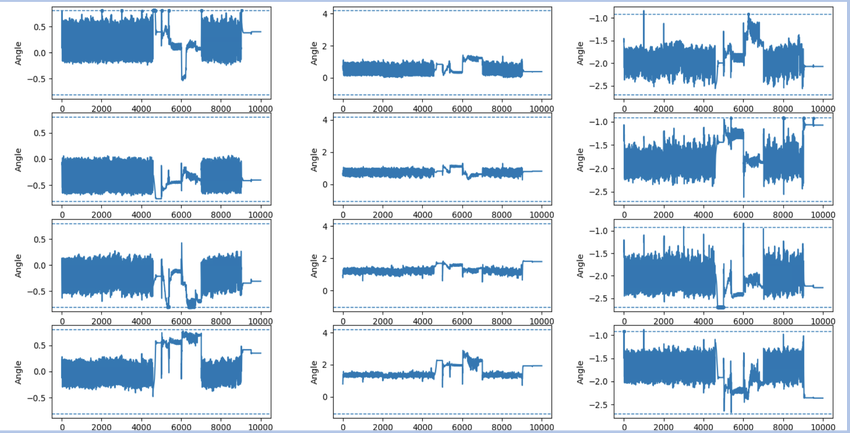}
            \caption{Softmin CBF}
            \label{fig:softmin_cbf}
        \end{subfigure}\hfill
        \begin{subfigure}[t]{0.49\linewidth}
            \centering
            \includegraphics[width=\linewidth, height=6cm, keepaspectratio=false]{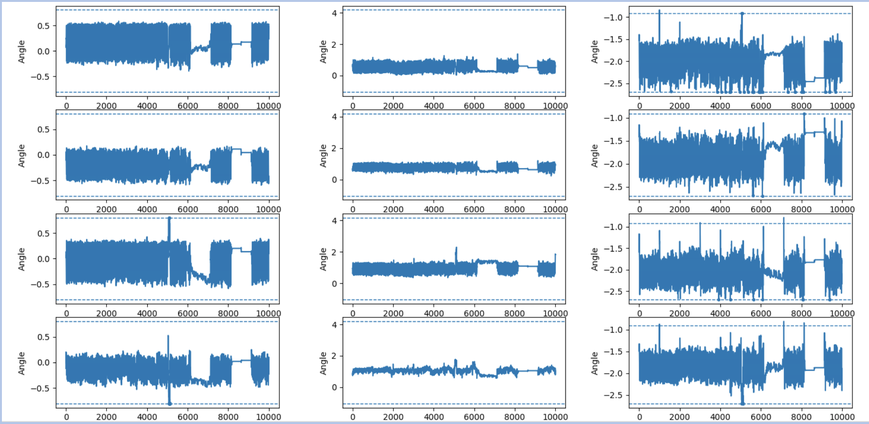}
            \caption{VOCBF}
            \label{fig:vocbf}
        \end{subfigure}
    \end{minipage}%
    }
    \caption{Joint angles of the 12 joints of the four reward-shaping strategies under the same conditions.}
    \label{fig:reward_comparison}
\end{figure*}
\begin{table}[t]
    \centering
    
    \label{tab:reward}
    \begin{tabular}{lcc}
        \toprule
        Reward & Equation & Weight \\
        \midrule
        Linear velocity tracking & $e^{-\frac{{||v^{cmd}_{xy}-v_{xy}||^2_2}}{0.25}}$ & 1\\
        Angular velocity tracking & $e^{\frac{{-||\omega^{cmd}_{yaw}-\omega_{yaw}||^2_2}}{0.25}}$ & 0.5\\
        Linear velocity ($z)$ & $v_z^2$ & -2\\
        Angular velocity ($xy)$ & $||\omega _{xy}||^2_2$ & -0.05 \\
        Orientation & $||g||^2_2$ & -0.2\\
        Joint power & $\dot{\theta}^T\tau$ & -0.00002\\
        Joint accelerations & $||\Ddot{\theta}||^2$ & -0.000000025\\
        Body height & $(h_{target}-h)^2$ & -1\\
        Foot clearance & $\sum_{i=0}^3(p_z^{target}-p_z^i)^2v^i_{xy}$ & -0.01\\
        Action rate & $||a_t-a_{t-1}||^2_2$ & -0.01\\
        Smoothness & $||a_t-2a_{t-1}+a_{t-2}||^2_2$ & -0.01\\
        \bottomrule
    \end{tabular}
    \caption{Reward from HIMLOCO same as performance}

\end{table}
\begin{table}[t]
    \centering
    
    \begin{tabular}{lcc}
        \toprule
        Reward & Safety violation (\%) $\downarrow$ & Performance $\uparrow$ \\
        \midrule
        Only clipping           & 27  & 168 \\
        Sum CBF (baseline)      & 14  & 163 \\
        Softmin CBF             & 2   & \textbf{170} \\
        Value function CBF      & \textbf{0.1} & 160 \\
        \bottomrule
    \end{tabular}
    \caption{Comparison of reward-shaping strategies. Lower safety violation and higher performance are better; best values are in bold.}\label{tab:compare}
\end{table}
From Fig.\ref{fig:barrier plot}, the learnt VOCBF barrier function is more conservative than the original barrier function as it accounts for the input constraints. It can be observed that for $x<\delta$, the VOCBF barrier function $\psi(x)>0$, this is because the joint angle 1 in quadruped doesn't reach these joint angles while demonstrations as well as during walking, similar features are observed in all the other 11 joints.

Testing was done  by giving random velocity commands every 100 timesteps for a total of 10000 time steps. The environment had various terrains ranging from plain floor to stairs and stairs. If the environment was simple, then even clipping doesn't have any safety violation, safety violation occurs when either the robot is at the steps or velocity is high. This the same reason why HIMLOCO is used as it learns the type of terrain using the past observations and provides the control input that maximizes the reward in the corresponding terrain, thus it provides a working policy in any general terrain.

From Table \ref{tab:compare}, it can be observed that safety violation is the least in case of VOCBF based reward shaping, it decreases from $27\%$ while clipping to $0.1\%$, even the next best, Softmin CBF give violation for around $2\%$ of the time, so the difference between violations are significant. In case of performance, there is no such clear difference between the methods as all the methods give performance in the range $160- 170$.

% In the preamble: \usepackage{subcaption}  (and \usepackage{graphicx})

% Preamble: \usepackage{graphicx}, \usepackage{subcaption}

In fig. \ref{fig:reward_comparison}, the dotted lines are the limits within which we want the angles to be in, it can be clearly observed that, in VOCBF method the joint angles are almost always within the required range whereas in other methods it reaches the limits more frequently.

% In the preamble: \usepackage{booktabs}

\section{Conclusions}
\label{section: Conclusions}
In this paper, we proposed a method using Value guided offline control barrier functions for reward shaping to enforce safety of the type $x_{min} \leq x \leq x_{max}$ in dynamical systems. This method doesn't need the dynamics to be known beforehand and even the VOCBF is learnt offline which makes the method simple. Experiment was done in Isaacgym's Unitree -Go1 environment and VOCBF based reward shaping had better safety and performance was similar to the other baselines- just using clipping, sum cbf and softmin cbf.

\label{section: References}
\bibliographystyle{IEEEtran}
\bibliography{references.bib}

@misc{tayal2026vocbflearningsafetyfilters,
      title={V-OCBF: Learning Safety Filters from Offline Data via Value-Guided Offline Control Barrier Functions}, 
      author={Mumuksh Tayal and Manan Tayal and Aditya Singh and Shishir Kolathaya and Ravi Prakash},
      year={2026},
      eprint={2512.10822},
      archivePrefix={arXiv},
      primaryClass={cs.AI},
      url={https://arxiv.org/abs/2512.10822}, 
}

@INPROCEEDINGS{9561949,
  author={Bansal, Somil and Tomlin, Claire J.},
  booktitle={2021 IEEE International Conference on Robotics and Automation (ICRA)}, 
  title={DeepReach: A Deep Learning Approach to High-Dimensional Reachability}, 
  year={2021},
  volume={},
  number={},
  pages={1817-1824},
  doi={10.1109/ICRA48506.2021.9561949}}

@misc{long2024hybridinternalmodellearning,
      title={Hybrid Internal Model: Learning Agile Legged Locomotion with Simulated Robot Response}, 
      author={Junfeng Long and Zirui Wang and Quanyi Li and Jiawei Gao and Liu Cao and Jiangmiao Pang},
      year={2024},
      eprint={2312.11460},
      archivePrefix={arXiv},
      primaryClass={cs.RO},
      url={https://arxiv.org/abs/2312.11460}, 
}

\end{document}